\pdfoutput=1

\documentclass[11pt]{article}

\usepackage{emnlp2021}

\usepackage{times}
\usepackage{latexsym}
\usepackage{graphicx}

\usepackage{comment}
\usepackage{booktabs}
\usepackage{multirow}
\usepackage{xcolor}

\usepackage{setspace}

\usepackage[T1]{fontenc}

\usepackage[utf8]{inputenc}

\usepackage{microtype}
\usepackage{pythonhighlight}

\lstnewenvironment{mypython}[1][]{\lstset{style=mypython,#1}}{}

\title{TweetNLP: Cutting-Edge Natural Language Processing for Social Media}

\author{
\bf{Jose Camacho-Collados}$^1$ ~ \bf{Kiamehr Rezaee}$^1$ ~ 
\bf{Talayeh Riahi}$^1$ ~ \bf{Asahi Ushio}$^1$ \\ \bf{Daniel Loureiro}$^1$ ~ \bf{Dimosthenis Antypas}$^1$ ~  \bf{Joanne Boisson}$^{1,6}$ ~ \bf{Luis Espinosa-Anke}$^{1,6}$ \\ \bf{Fangyu Liu}$^2$ ~ \bf{Eugenio Martínez-Cámara}$^3$ ~ \bf{Gonzalo Medina}$^3$ \\ \bf{Thomas Buhrmann}$^4$ ~  \bf{Leonardo Neves}$^5$ ~ \bf{Francesco Barbieri}$^5$ \\
$^1$Cardiff NLP, Cardiff University, UK ~
$^2$LTL, University of Cambridge, UK \\
$^3$DaSCI, University of Granada, Spain ~
$^4$Graphext, Spain ~
$^5$Snap Inc., USA ~ 
$^6$AMPLYFI, UK ~ \\
{\tt cardiffnlp.contact@gmail.com}
}

\begin{document}
\maketitle


\begin{abstract}
In this paper we present TweetNLP, an integrated platform for Natural Language Processing (NLP) in social media. TweetNLP supports a diverse set of NLP tasks, including generic focus areas such as sentiment analysis and named entity recognition, as well as social media-specific tasks such as emoji prediction and offensive language identification. Task-specific systems are powered by reasonably-sized Transformer-based language models specialized on social media text (in particular, Twitter) which can be run without the need for dedicated hardware or cloud services. The main contributions of TweetNLP are: (1) an integrated Python library for a modern toolkit supporting social media analysis using our various task-specific models adapted to the social domain; (2) an interactive online demo for codeless experimentation using our models; and (3) a tutorial covering a wide variety of typical social media applications. 
\end{abstract}



\section{Introduction}

Today's society cannot be understood without the role of social media.
Online users connect more and more via platforms that enable content sharing, either generic or around specific topics, and do this by means of text-only messages, or augmenting them with multimedia content such as pictures, audio or video. 
As such, these platforms have been used to understand user, group and organization-wide behaviours \cite{Yang_Wang_Pierce_Vaish_Shi_Shah_2021,hu2021rise}. In particular, Twitter, which is the main platform studied in this paper, has long been an important resource for understanding society at large \cite{weller2013twitter}.
Twitter is interesting for NLP because it embodies many features that are natural in spontaneous and ever-evolving fast-paced communication.
However, the majority of NLP research focuses on optimizing model development against training data and evaluation benchmarks which are, at worst, reasonably clean (e.g., news articles, blog posts or Wikipedia).
Consequently, when deployed \textit{in the wild}, features such as noisiness, multilinguality, immediacy, slang, technical jargon, lack of context, platform-specific restrictions on message length, emoji and other modalities, etc. become core communicative variables that need to be factored in.
Indeed, even traditional NLP tasks such as normalization \cite{han2011lexical,baldwin2015shared}, POS tagging \cite{derczynski-etal-2013-twitter}, sentiment analysis \cite{poria2020beneath} or named entity recognition \cite{ritter2011named,baldwin-etal-2013-noisy} have been shown to produce suboptimal results in the context of social media.

Given the above, we put forward TweetNLP 
(\url{tweetnlp.org}), which offers a full-fledged NLP platform specialized in Twitter. The backbone of TweetNLP consists of Transformer-based language models that have been trained on Twitter \cite{barbieri-etal-2020-tweeteval,barbieri-espinosaanke-camachocollados:2022:LREC,loureiro-etal-2022-timelms}. Then, these specialized language models have been further fine-tuned for specific NLP tasks on Twitter data. 
These models have already proved highly popular, with thousands of downloads from the Hugging Face model hub every month \cite{wolf-etal-2020-transformers}.\footnote{Most notably, the sentiment analysis model has been the most downloaded model in the Hugging Face model hub in January 2022, with over 15M downloads. Similarly, the TweetEval benchmark, in which most task-specific Twitter models are fine-tuned, has been the second most downloaded dataset in April 2022, with over 150K downloads.} 
TweetNLP integrates all these resources into a single platform. With a simple Python API, TweetNLP offers an easy-to-use way to leverage cutting-edge NLP models in a variety of social media tasks.
Despite the trend of ever-larger language models \cite{shoeybi2019megatron,brown2020language}, TweetNLP is more focused on the general user and applicability, and therefore integrates base models which are easily run in standard computers or on free cloud services.
Finally, all models can be accessed from an interactive online demo, offering anyone the possibility to test models and perform real-time analysis on Twitter.

\section{Related Work}


General-purpose NLP libraries have been available for many years. Starting from the Java-based CoreNLP \cite{manning-EtAl:2014:P14-5} to the more recent Python-based library Stanza \cite{qi2020stanza}. More recently, libraries such as spaCy\footnote{\url{https://spacy.io}}
have been ubiquitous in NLP, both in research and industry. Finally, in the language models and Transformers era, the Hugging Face Transformer hub has become indispensable for state-of-the-art NLP \cite{wolf-etal-2020-transformers}, which is also leveraged for our library TweetNLP. However, none of these libraries is specialized in social media or Twitter.

As for libraries developed specifically for social media, these are more limited and mostly associated with low-level tasks such as tokenization, part-of-speech \cite{owoputi-etal-2013-improved} tagging or dependency parsing \cite{kong-etal-2014-dependency}, and initially available for Java.
The most recent Twitter-specific Python library is TweebankNLP \cite{jiang-EtAl:2022:LREC2} based on Stanza. This library provides state-of-the-art models on tokenization and lemmatization, besides competitive models on NER, part-of-speech tagging and dependency parsing.
In contrast, TweetNLP focuses on specialized Twitter-specific language models for downstream tasks such as sentiment analysis and offensive language identification.


\section{Models and Functionalities}

TweetNLP is versatile in that it covers a wide range of tasks and applications. The backbone of TweetNLP are transformer-based language models, which are covered in Section \ref{languagemodels}. The concrete NLP tasks integrated in TweetNLP are presented in Section \ref{tasks}. Finally, in Section \ref{embeddings} we present embeddings used to represent words and tweets. All TweetNLP model checkpoints are available in the appendix.




\subsection{Language models}
\label{languagemodels}

Language models are at the core of TweetNLP. 
Instead of relying on general-purpose models \cite{devlin-etal-2019-bert} or training a language model from scratch \cite{nguyen2020bertweet}, we start from RoBERTa \cite{liu2019roberta} and XLM-R \cite{conneau-etal-2020-unsupervised} checkpoints and continue pre-training on Twitter-specific corpora. This was shown to be generally more reliable for the amount of text analysed in \newcite{barbieri-etal-2020-tweeteval}. 
Given our aim for democratizing the usage of specialized language models for social media, another important feature of TweetNLP is the relatively small size of the language models. To this end, all language models rely on the equivalent of a RoBERTa-base or XLM-R-base architecture.
These models are efficient on standard hardware and free-tiers of cloud computing services, with reasonable speed even without GPU support.



\paragraph{TweetEval \cite{barbieri-etal-2020-tweeteval}.} This model was initially released as part of the TweetEval project. It is based on a RoBERTa-base architecture using the original model as an initial check point \cite{liu2019roberta}. Later, this language model was further pre-trained on a corpus of 60M tweets from May 2018 to August 2019.

\paragraph{TimeLMs \cite{loureiro-etal-2022-timelms}.} This model is initially trained on the same Twitter corpus used by \citet{barbieri-etal-2020-tweeteval}. The main difference lies on a few preprocessing improvements applied to the underlying corpus, including measures to reduce potential spam and near duplicates, and more recent corpora used for continual pretraining. 
The overall quantity of tweets is therefore larger, as the model is regularly updated (every 3 months) with a fixed number of additional tweets. 
The most recently released TimeLMs model to date is pre-trained on 132M tweets until the end of June 2022.


\paragraph{XLM-T \cite{barbieri-espinosaanke-camachocollados:2022:LREC}.} This model was trained on 198M tweets on over thirty languages from May 2018 to March 2020, following a similar strategy to \citet{barbieri-etal-2020-tweeteval}. In this case, the initial checkpoint was XLM-R-base \cite{conneau-etal-2020-unsupervised}.

\subsection{Supported tasks}
\label{tasks}

In the following we describe the tasks supported by TweetNLP. For the tweet classification tasks included in TweetEval, and for topic classification, we simply fine-tune the models described above on the corresponding datasets, as described in \newcite{barbieri-etal-2020-tweeteval}. For model fine-tuning on named entity recognition, we rely on the T-NER library \cite{ushio-camacho-collados-2021-ner}, which is also integrated into TweetNLP.

\paragraph{Sentiment Analysis.} 
The sentiment analysis task integrated in TweetNLP 
consists of predicting the sentiment of a tweet with one of the three following labels: positive, neutral or negative.
The base dataset for English is the unified TweetEval version of the Semeval-2017 dataset from the task on \textit{Sentiment Analysis in Twitter} \cite{rosenthal2017semeval}. Moreover, for the languages other than English we include the datasets integrated in UMSAB \cite{barbieri-espinosaanke-camachocollados:2022:LREC}, namely Arabic \cite{rosenthal2017semeval}, French \cite{benamara2017analyse}, German \cite{cieliebak2017twitter}, Hindi \cite{patra2015shared}, Italian \cite{barbieri2016overview}, Portuguese \cite{brum2017building}, and Spanish \cite{diaz2018democratization}.  

\paragraph{Emotion Recognition.}
Given a tweet, this task consists of associating it with its most appropriate emotion. As a reference dataset we use the SemEval 2018 task on \textit{Affect in Tweets}  \cite{mohammad2018semeval}, simplified to only the four emotions used in TweetEval: anger, joy, sadness and optimism. 

\paragraph{Emoji Prediction.}
The goal of emoji prediction is to predict the final emoji on a given tweet. The dataset used to fine-tune our models is the TweetEval adaptation from the SemEval 2018 task on \textit{Emoji Prediction} \cite{barbieri2018semeval}, including 20 emoji as labels. 

\paragraph{Irony Detection.} This is a binary classification task that aims at detecting whether a tweet is ironic or not. It is based on the \textit{Irony Detection} dataset from the SemEval 2018 task \cite{van2018semeval}. 

\paragraph{Hate Speech Detection.} The hate speech dataset consists of detecting whether a tweet is hateful towards women or immigrants. It is based on the \textit{Detection of Hate Speech} task at SemEval 2019 \cite{basile-etal-2019-semeval}. 

\paragraph{Offensive Language Identification.} The task consist of identifying any form of offensive language in a tweet. The dataset is based on the SemEval 2019 task on \textit{Identifying and Categorizing Offensive Language in Social Media} \cite{zampieri-etal-2019-semeval}. 

\paragraph{Stance Detection.} Given a target topic and a tweet, stance detection consists of assessing whether the author of the tweet has a positive, neutral or negative position towards the target. The dataset considered was initially released for the SemEval 2016 task on \textit{Detecting Stance in Tweets} \cite{mohammad2016semeval}. 

\paragraph{Topic Classification.} The aim of this task is, given a tweet, assign topics related to its content. The task is formulated as a supervised multi-label classification problem where each tweet is assigned one or more topics from a total of 19 available topics. The topics were carefully curated based on Twitter trends with the aim to be broad and general, consisting of classes such as: \textit{arts and culture}, \textit{music}, or \textit{sports}. The underlying tweet topic classification dataset contains over 10K manually-labeled tweets \cite{antypas2022twitter}. 

\paragraph{Named Entity Recognition.} 
The goal of named entity recognition (NER) is to find entities and identify their entity types in a given sentence. 
The underlying Twitter NER dataset is composed of over 10K tweets which were annotated (internally) with seven entity types.\footnote{More details about the datasets for topic classification and named entity recognition will be provided at a later stage. Datasets were annotated internally in Snap and we are working on releasing them to the public according to regulations.}



\subsection{Embeddings}
\label{embeddings}

In addition to the language models and their supported tasks, we also release high-quality vector representation models for words and tweets, i.e., \textit{embeddings} \cite{pilehvar2020embeddings}. These relatively low-dimensional vector representations can be exploited for a different range of applications and analyses such as word/tweet similarity or tweet retrieval, to name a few.


\paragraph{Word embeddings.} TweetNLP word embeddings are based on fastText \cite{bojanowski2017enriching} and trained on the same corpora used to train the language models described in Section \ref{languagemodels}.
In particular, we trained two sets of embeddings: (1) a monolingual English model trained with the TimeLMs Twitter corpus until the end of 2021; and (2) a multilingual model trained with the Twitter corpus used for XLM-T.
Both models were trained using the official fastText package with 300 dimensions, minimum n-gram size 2, maximum n-gram size 12, and remaining parameters set to defaults.




\paragraph{Tweet embeddings.} For tweet embeddings, we pulled tweet-reply pairs from the Twitter API and trained contrastive embeddings with an InfoNCE loss \citep{oord2018representation}. For tweets with multiple replies, we randomly sampled one reply. In training, one mini-batch is composed of a list of tweet-reply pairs. The tweet-reply pairs are regarded as positive samples; the enumeration of all other possible combination of tweet-reply, tweet-tweet, and tweet-reply pairs are regarded as negative samples. The contrastive InfoNCE loss then pulls positive pair representations close while pushes negative representations away from each other.
Training was performed on 1.1M tweet-reply pairs, and we collected a separate tweet-reply set of 10k pairs for selecting the model checkpoint.




\section{TweetNLP Python library}
\label{pythonlibrary}

The TweetNLP Python library has been integrated into pypi\footnote{\url{https://pypi.org/project/tweetnlp/}} and therefore is easily accessible and can be installed from pip ("pip install tweetnlp"). All the details on how to use TweetNLP are in the associated Github repository, which is released fully open-source: \url{https://github.com/cardiffnlp/tweetnlp}.


Once installed, loading and using a fine-tuned model on any specific task can be done as follows.


\begin{mypython}[label=SO-test]
    from tweetnlp import load
    tweet = "I love Paris!!"
    # Sentiment Analysis
    model = load('sentiment')
    model.sentiment(tweet) 
    # Tweet Embeddings
    model = load('sentence_embedding')
    model.embedding(tweet)
    # Masked Language Model
    model = load('language_model')
    tweet = "I love <mask>!!"
    model.mask_prediction(tweet)
\end{mypython} 


\noindent With the \textit{load} statement, the associated fine-tuned language models are loaded in the background. Users can then get the predictions for any given sentence or tweet with a simple pre-defined function (e.g., \textit{.sentiment} or \textit{.predict}). Custom loading of existing fine-tuned language models not included in TweetNLP is also possible.  
The same functionalities apply to all the other tasks described in Section \ref{tasks}.


\section{Tutorials}
\label{tutorials}

In addition to the Python library presented in the previous section, TweetNLP offers access to the underlying Python code structured in instructive Google Colab notebooks with starter code and examples (\url{https://tweetnlp.org/get-started/}). These notebooks are aimed at users with varying degrees of experience in NLP and social media processing. In the following we list the currently existing tutorials and a brief description:



\paragraph{Introduction to TweetNLP.} In this initial introduction, users learn how to use the TweetNLP Python library to make use of specialized models in social media for a wide variety of tasks from sentiment analysis to named entity recognition. 

\paragraph{Getting data from Twitter.} This notebook helps users understand the Twitter API\footnote{\url{https://developer.twitter.com/en/docs/twitter-api}} and how to interact with it. More importantly, there are concrete examples on how to retrieve data (i.e. tweets) from Twitter, usually given a hashtag or a keyword.


\paragraph{Custom fine-tuning.} In this notebook users can learn to fine-tune any given language model on a specific task (e.g. sentiment analysis). For this, we will take advantage of the TweetEval task data and unified format \cite{barbieri-etal-2020-tweeteval}. Additionally, users can learn how to easily evaluate language models on TweetEval. 

\paragraph{Word embeddings.} With this notebook users can learn how to train their own word embeddings on custom data using Gensim\footnote{\url{https://radimrehurek.com/gensim/}} \cite{Rehurek_Software_Framework_for_2010}. The notebook also includes examples on how to get similarity scores from Twitter-specific word embeddings, or how to obtain the nearest neighbour words from a given input word.

\paragraph{Language models over time.} This notebook leverages the TimeLMs library \cite{loureiro-etal-2022-timelms}. Users can learn how to make use of language models that have been trained in short periods of time since 2019 until recently.

\paragraph{Tweet embeddings.} This notebook contains examples on how to transform a tweet into a vector (embedding) and how these enable important applications such as tweet similarity and retrieval.

\section{Demo}

\begin{figure}[h]
 \centering
 \includegraphics[width=1.0\columnwidth]{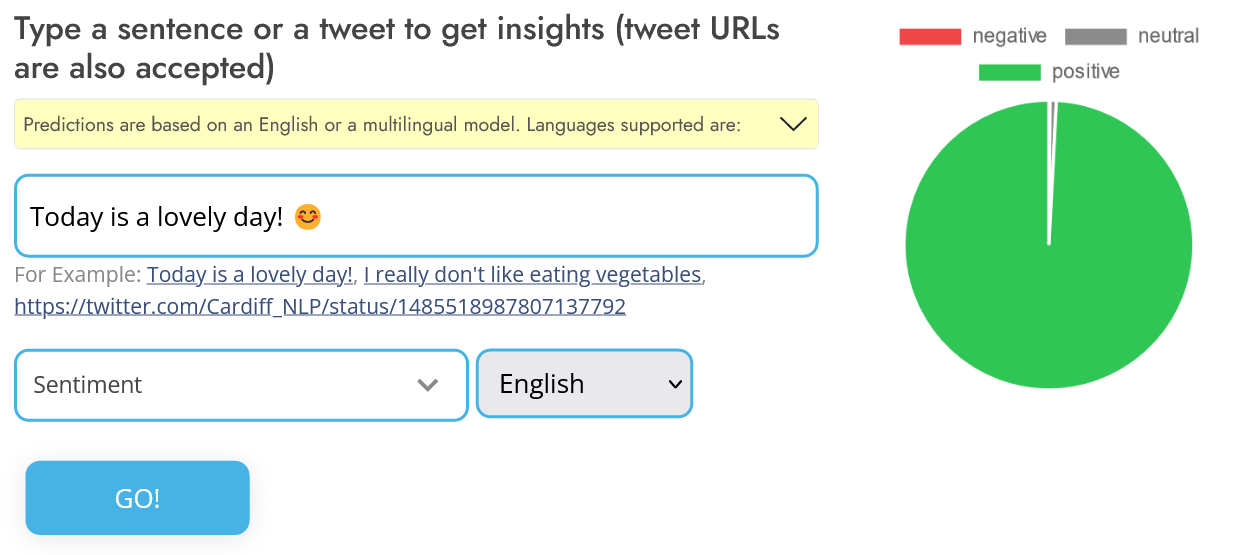}
\caption{TweetNLP tweet classification demo.}
 \label{fig:sentimentdemo}
\end{figure}

In addition to the Python-based library and tutorials, we developed a comprehensive web-based demo integrating all our models, available at \url{https://tweetnlp.org/demo/}. The goal of the demo is for any user to be able to test our models and get predictions. In particular, the model includes the following five functionalities:



\paragraph{Sentence/tweet classification (Figure \ref{fig:sentimentdemo}).} 
Users can input a sentence or a tweet (including a tweet URL) and the output is a plot display of the confidence of the model with respect to its predictions. This demo includes all tweet classification tasks supported in English (see Section \ref{tasks}), as well as a multilingual sentiment analysis model based on XLM-T. 

\begin{figure}[h]
 \centering
 \includegraphics[width=1.0\columnwidth]{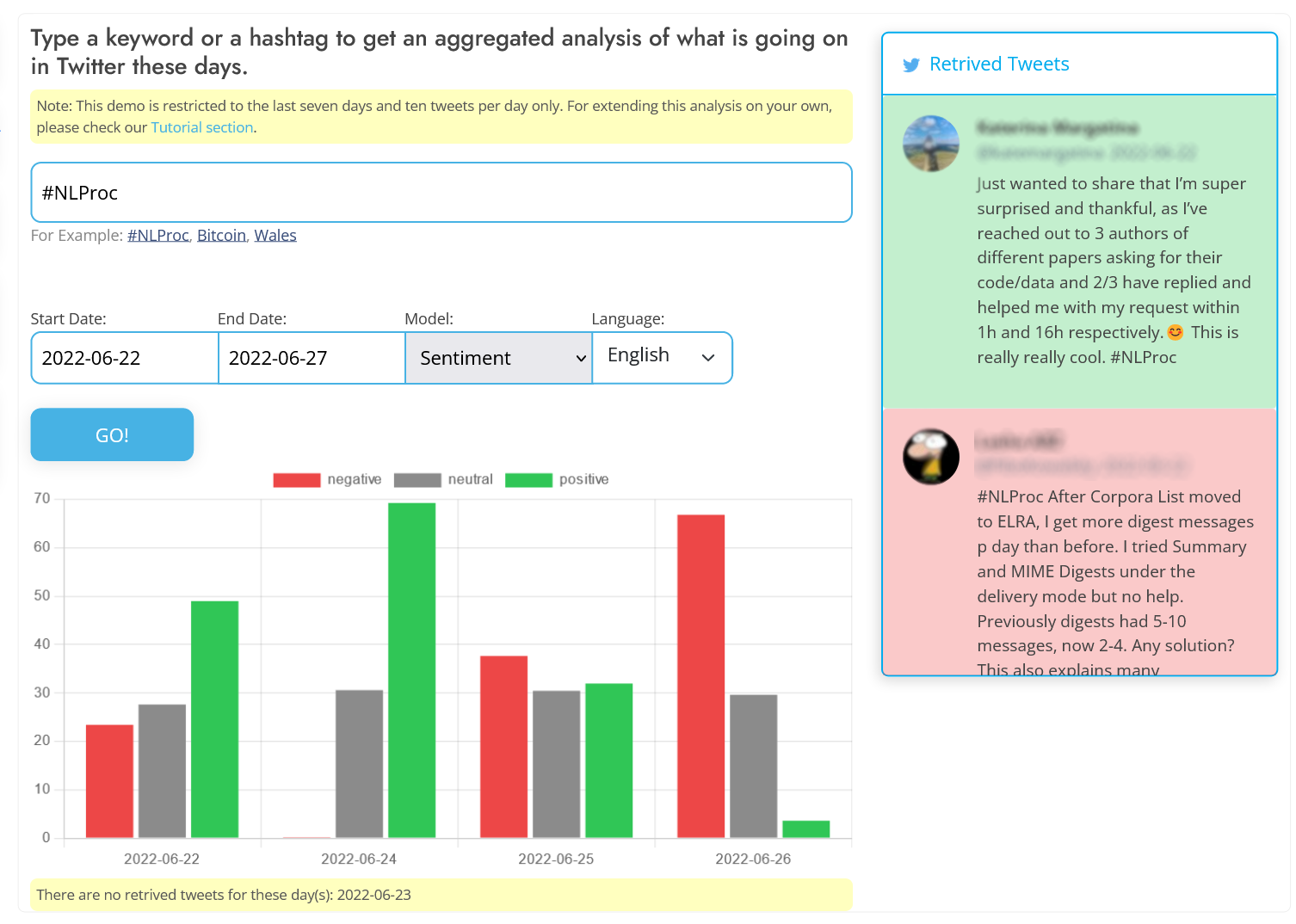}
\caption{TweetNLP hashtag analysis demo. The output is a bar plot that shows the sentiment of the retrieved tweets over time for the input hashtag \#NLProc.} 
 \label{fig:hashtagdemo}
\end{figure}

\paragraph{Hashtag analysis (Figure \ref{fig:hashtagdemo}).} 
This demo directly interacts with the Twitter API. Users can type a hashtag (or any keyword), initial and end dates, task and language. The system will then retrieve tweets for the given time interval and compute an aggregated analysis of the results. Languages supported for this demo are available in the appendix.

\begin{figure}[h]
 \centering
 \includegraphics[width=1.0\columnwidth]{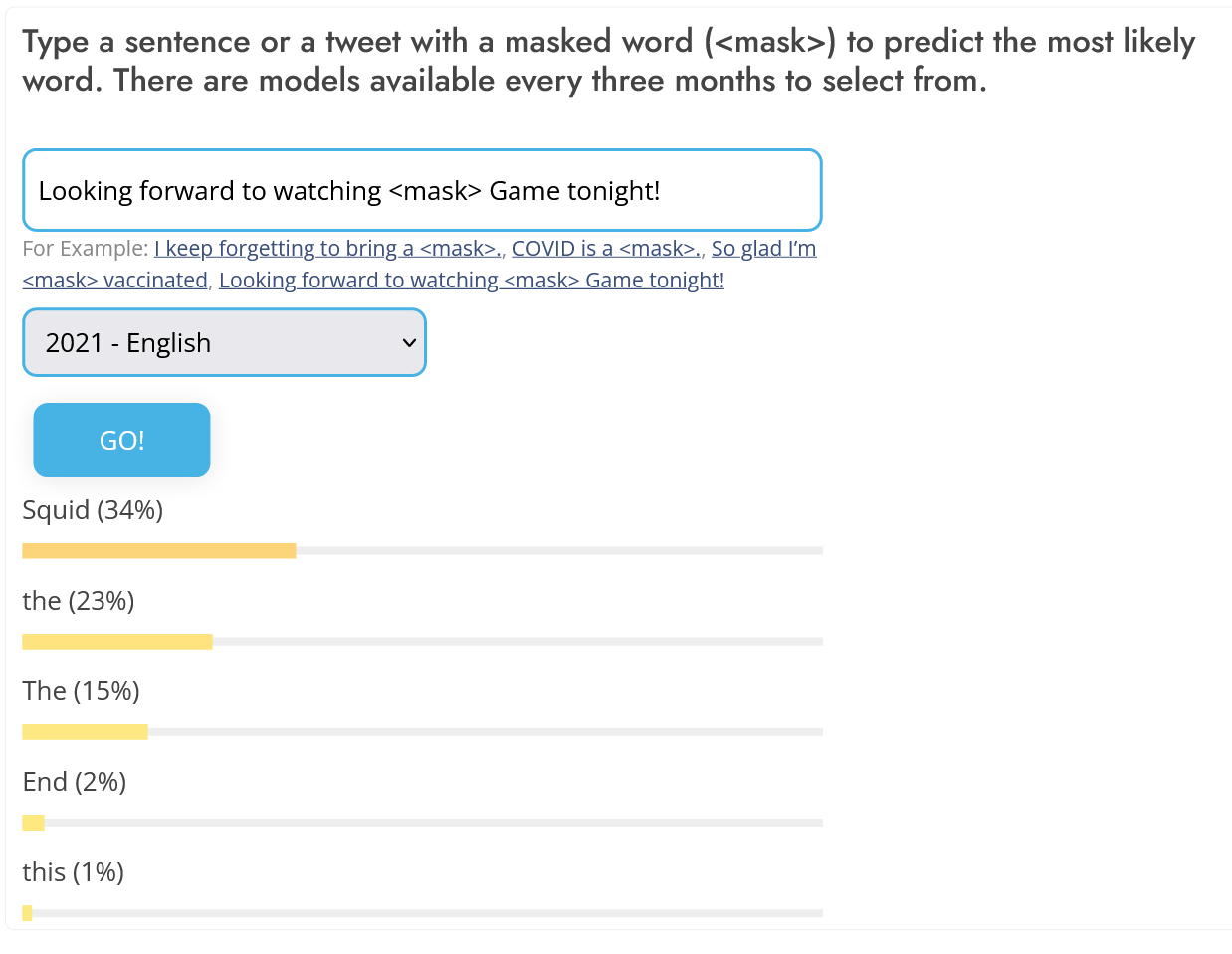}
\caption{TweetNLP word prediction demo.}
 \label{fig:timelmsdemo}
\end{figure}

\paragraph{Word prediction (Figure \ref{fig:timelmsdemo}).} 
Masked language models utilized in TweetNLP are trained to predict unknown (or \textit{masked}) words within a sentence. For this demo, users can input a sentence with a masked word and the system will show the most likely words as given by the masked language model, in order of confidence.  

\begin{figure}[ht]
 \centering
 \includegraphics[width=1.0\columnwidth]{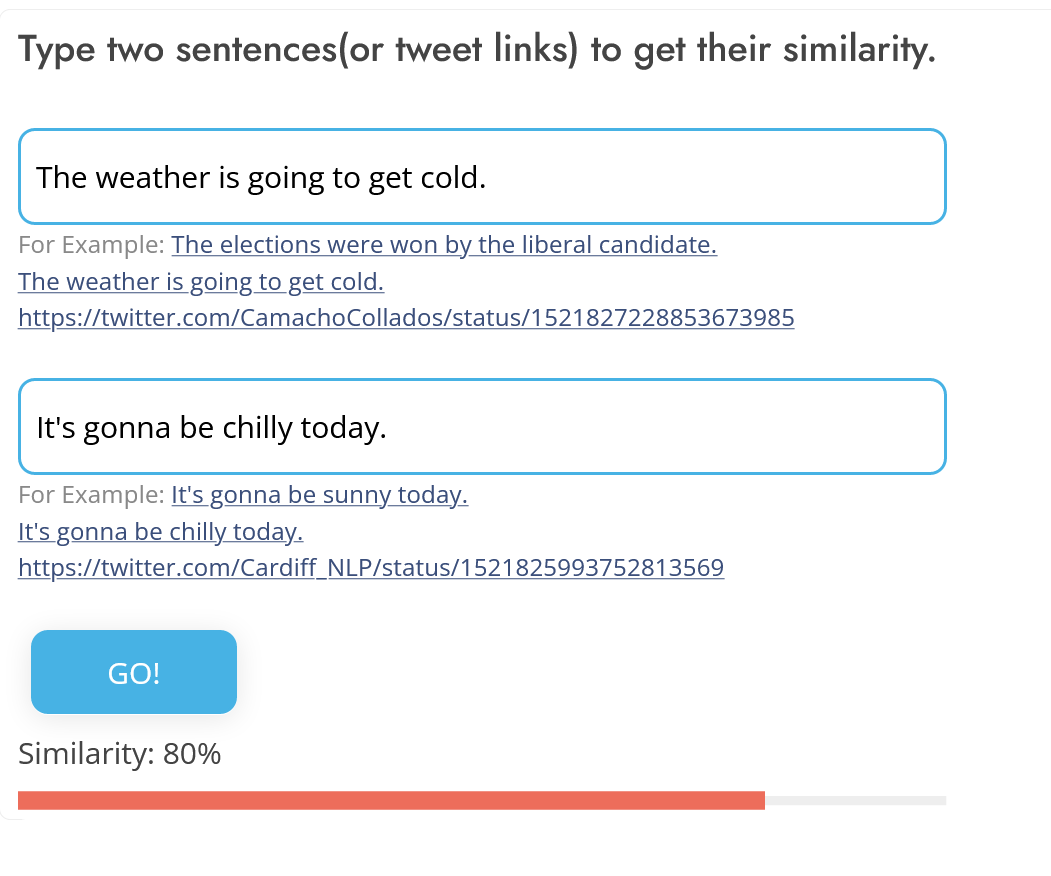}
\caption{TweetNLP tweet similarity demo.}
 \label{fig:tweetembeddingsdemo}
\end{figure}

\paragraph{Tweet similarity (Figure \ref{fig:tweetembeddingsdemo}).} Given two short pieces of text (e.g., two sentences or two tweets), this demo displays their cosine similarity score on a 0-100 scale as provided by our default tweet embedding model. 

\begin{figure}[h]
 \centering
 \includegraphics[width=1.0\columnwidth]{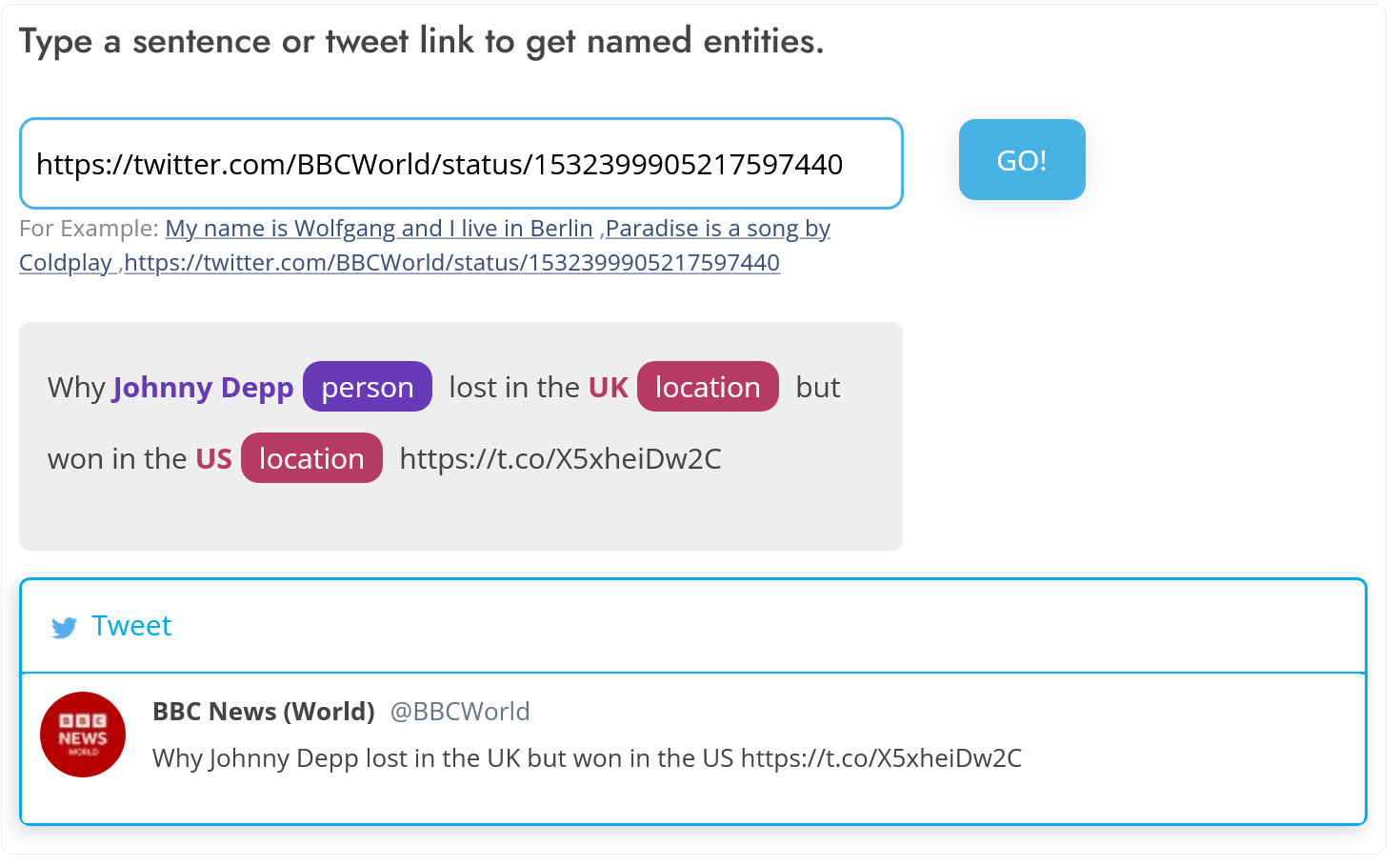}
\caption{TweetNLP Named Entity Recognition demo.}
 \label{fig:nerdemo}
\end{figure}

\paragraph{Named Entity Recognition (Figure \ref{fig:nerdemo}).} Given a tweet or a sentence, this NER demo locates its named entities and infers their types.

\section{Evaluation}
\label{sec:eval}

In this section, we provide experimental results of the default models integrated into TweetNLP. 

\subsection{Experimental setting}


\paragraph{Datasets.} For the evaluation we utilized all the train/validation/test splits described in Section \ref{tasks}. In particular, we relied on the TweetEval-released datasets for all tweet classification tasks except for topic classification. 


\paragraph{Default TweetNLP language models.} While in TweetNLP all Twitter-specific language models are included, we use as a default (1) TimeLMs trained until December 2021 for English and (2) XLM-T for the languages other than English and multilingual tasks. These models are then fine-tuned to the corresponding tasks as described in Section \ref{tasks}.

\paragraph{Comparison systems.}
We report the performance of all original TweetEval baselines \cite{barbieri-etal-2020-tweeteval}: a frequency-based SVM classifier, fastText \cite{joulin-etal-2017-bag}, a Bidirectional LSTM, RoBERTa-base \cite{liu2019roberta}, a RoBERTa-base model trained on Twitter from scratch (RoB-Twitter) and the original TweetEval RoBERTa-base model. As another baseline we include BERTweet \cite{nguyen2020bertweet}, trained on almost 1 billion tweets from 2013 to 2019. 

\paragraph{Language model fine-tuning.} Fine-tuning is performed on the training sets of each corresponding dataset, using their corresponding development sets for validation. We followed TweetEval training protocols for tweet classification, where only the learning rate and number of epochs are tuned \cite{barbieri-etal-2020-tweeteval}. All reported results for language models are based on an average of three runs.

\subsection{Results}


\begin{table*} 
\centering
\resizebox{\textwidth}{!}{
\begin{tabular}{c|c|c|c|c|c|c|c|c|c}
\toprule
   &
  \textbf{Emoji} &
  \textbf{Emotion} &
  \textbf{Hate} &
  \textbf{Irony} &
  \textbf{Offensive} &
  \textbf{Sentiment} &
  \textbf{Stance} &
  \textbf{Topic}
  &
  \textbf{NER}\\ \hline 

  SVM &
  29.3 &
  64.7 &
  36.7 &
  61.7 &
  52.3 &
  62.9 &
  67.3 &
  30.5 &
  - \\
  fastText &
  25.8 &
  65.2 &
  50.6 &
  63.1 &
  73.4 &
  62.9 &
  65.4 &
  24.0 &
  - \\

  BLSTM &
  24.7 & 66.0 & 52.6 & 62.8 & 71.7 & 58.3 & 59.4 &
  27.0 &
  - \\
  RoB-Base &
  30.9 & 76.1 &
  46.6 & 59.7 & 79.5 & 71.3 &
  68.0 & 50.1 & 58.0
    \\ \cline{1-10}

  RoB-Twitter &
  29.3 &
  72.0 & 46.9 & \textbf{65.4} &
  77.1 &
  69.1 &
  66.7 &
  - & - \\  
  
     TweetEval &
  31.4 &
  78.5 &
  52.3 &
  61.7 &
  80.5 &
  72.6 &
  69.3 &
  56.8 & 56.8
     \\  
  
  BERTweet &
  33.4 &
  79.3 &
  \textbf{56.4} & 
  \textit{~82.1*} &
  79.5 & 73.4 & 71.2 &
  52.7 & 58.7 \\
    
      TweetNLP (TimeLMs-21) &
    \textbf{34.0} &
  \textbf{80.2} &
  55.1 &
  64.5 &
  \textbf{82.2} &
  \textbf{73.7} &
  \textbf{72.9} & \textbf{58.8} & \textbf{59.7}
  
  
  

   \\ \hline \hline
\multicolumn{1}{c|}{Evaluation metric} &
  M-F1 &
  M-F1 &
  M-F1 &
  F$^{(i)}$ &
  M-F1 &
  M-Rec &
  AVG (F) & M-F1 & M-F1
   \\ \bottomrule
\end{tabular}
}
\caption{\label{table-results-english} Test results in the nine TweetNLP-supported tasks.}
\end{table*}

Table \ref{table-results-english} shows the main results of our TweetNLP default language model and comparison systems on nine Twitter-based tasks.\footnote{The BERTweet result on Irony is marked with * as their pre-training corpus overlapped with the Irony dataset, which was constructed using distant supervision.} 
The default TimeLMs-21 model achieves the overall results on most tasks, especially comparing it with a comparable general-purpose RoBERTa-based model. In the following we also provide details of our experimental results on languages other than English , and for the integrated word and tweet embedding models.






\paragraph{Multilingual sentiment analysis results.} In addition to the English evaluation, we report results on multilingual sentiment analysis (Table \ref{table-results-multilingual}). The evaluation is performed on the UMSAB multingual sentiment analysis benchmark \cite{barbieri-espinosaanke-camachocollados:2022:LREC}. For this evaluation we compare XLM-T fine-tuned on all the language-specific training sets of UMSAB with XLM-R \cite{conneau-etal-2020-unsupervised} using the same fine-tuning strategy. As an additional indicative baseline, we include fastText trained on the language-specific training sets. As can be observed, our domain-specific XLM-T language model achieves the best overall results in all languages, further reinforcing the importance of in-domain language model training.

\begin{table*} 
\centering
\resizebox{\textwidth}{!}{
\begin{tabular}{c|c|c|c|c|c|c|c|c||c}
\toprule
   &
  \textbf{Arabic} &
  \textbf{English} &
  \textbf{French} &
  \textbf{German} &
  \textbf{Hindi} &
  \textbf{Italian} &
  \textbf{Portuguese} &
  \textbf{Spanish} &
  \textbf{ALL}
\\ \hline 

  fastText & 45.98 & 50.85 & 54.82 & 59.56 & 37.08 & 54.65 & 55.05 & 50.06 & 51.01
  \\
  XLM-R & 64.31 & 68.52 & 70.52 & 72.84 & 53.39 & 68.62 & 69.79 & 66.03 & 66.75
  \\
  TweetNLP (XLM-T) & \textbf{66.89} &  \textbf{70.63} & \textbf{71.18} & \textbf{77.35} & \textbf{56.35} & \textbf{69.06} & \textbf{75.42} & \textbf{68.52} & \textbf{67.91}
  \\
 \bottomrule
\end{tabular}
}
\caption{\label{table-results-multilingual} Sentiment analysis results (Macro-F1) on the UMSAB unified benchmark. XLM-R and TweetNLP models are fine-tuned on the training sets of all languages.}
\end{table*}



\paragraph{Word embedding results.} As a sanity check to verify the quality of the word embeddings, we simply test them on standard word similarity datasets: The WS-Sim similarity and WS-Rel relatedness subsets \cite{Agirreetal:09} from WordSim-353 \cite{Levetal:2002}, SemEval-2017 \cite{camacho-collados-etal-2017-semeval} and MEN \cite{bruni2014multimodal}. Then, we compared the results with the pre-trained fastText model trained on the Common Crawl \cite{bojanowski2017enriching}, and Wikipedia. According to Spearman correlation, the results of our Twitter embeddings were 0.77 (WS-Sim), 0.72 (WS-Rel), 0.69 (SemEval), and 0.79 (MEN).\footnote{While not directly comparable given the different sizes, we also compared with our previously-released Twitter-specific 100-dimensional fastText embeddings \cite{camacho2020learning}. The results for these embeddings were consistently lower: 0.65 (WS-Sim), 0.43 (WS-Rel), 0.52 (SemEval), and 0.76 (MEN).} In contrast, the pre-trained fastText Common Crawl results were 0.84 (WS-Sim), 0.64 (WS-Rel), 0.67 (SemEval), and 0.81 (MEN). We should note that these datasets are not specific to social media and even so, our trained embeddings outperform the standard pre-trained fastText in two datasets. In particular, there seems to be a marked difference between similarity and relatedness, where our Twitter embeddings appear to be more suited to relatedness.

\paragraph{Tweet embedding results.} For tweet embeddings we explore a tweet retrieval task setting which consists of finding the reply to a given tweet from the 10k replies in the search space. We randomly sampled 3k tweet-reply pairs that do not overlap with training data and split them into 3 sets of 1k pairs. We report accuracy@1 and average models' performance on the 3 sets. We also include results on sentence similarity, using the STS-benchmark \cite{cer2017semeval} and reporting Spearman's correlation. We list tweet-reply retrieval accuracy and STS-benchmark Spearman's correlation in Table \ref{table:tweet_embeddings}. We compare with recent supervised \cite[Sentence-BERT; all-mpnet-base-v2]{reimers-gurevych-2019-sentence}, and unsupervised  \cite[Mirror-BERT]{liu-etal-2021-fast}, \cite[SimCSE]{gao-etal-2021-simcse} sentence embedding models.\footnote{Baseline checkpoint links are included in the Appendix.} 
On the task of tweet-reply retrieval, our tweet-embeddings model significantly outperforms all-mpnet-base-v2 
trained with around 1B sentence pairs. This highlights the importance of in-domain training. On the STS-Benchmark, all-mpnet-base-v2 achieves the best performance and our tweet-embeddings perform the worst among baselines but they are generally in a similar ballpark. To complement this evaluation, we plan to test our tweet embeddings with a textual similarity dataset in the tweet domain in the future.


\begin{table}
\small
\centering
\begin{tabular}{lcc}
\toprule
Model & Retrieval & STS\\
\midrule
Sentence-BERT & 6.1 & 77.0 \\
all-mpnet-base-v2 & 15.8 & \textbf{83.4} \\
Mirror-RoBERTa & 8.8 & 79.6 \\
SimCSE-RoBERTa  & 9.2 & 80.3 \\
\midrule
TweetNLP (Tweet-embeddings) & \textbf{26.7} & 70.7 \\
 \bottomrule
\end{tabular}
\caption{Results of sentence and tweet embedding models on tweet-reply retrieval and the STS-benchmark.
}
\label{table:tweet_embeddings}
\end{table}

\section{Conclusion and Future Work}

In this demo paper we have presented TweetNLP, an all-round platform for NLP specialized in social media. The platform is powered by relatively lightweight language models trained on Twitter, and adapted (fine-tuned) to various popular NLP tasks on social media, such as sentiment analysis and offensive language identification. In addition to sharing the models, TweetNLP provides an online demo, a Python library, and a tutorial to make the most of the models, regardless of the expertise of the user. TweetNLP also enables easy inspection of the models by non-programmers, which can help identify harmful biases or errors, that in turn would help improve the models in the future.

While this first release version of TweetNLP is self-contained and complete, our goal is to keep updating it with both new models and tasks.
Since social media data is at the core of TweetNLP, we are planning to develop new datasets and models for social media tasks.
In particular, our idea is to go beyond tweet classification tasks, which are currently well covered in TweetNLP. For instance, low-level tasks such as syntactic parsing and part-of-speech tagging has been traditionally hard in noisy environments such as social media. Finally, in the future we are also planning to extend TweetNLP to other social media platforms such as Reddit, LinkedIn or Instagram, and to provide support for languages other than English in a wider variety of tasks.

\section{Impact Statement}

This paper deals with social media data, in particular with Twitter. All Twitter regulations were followed and data was extracted through the official Twitter API. To mitigate the potential effect of working with this type of data, all dataset-related tweets were anonymized, with URLs removed. In most cases dataset creators made an effort to remove offensive or harmful content from the tweets. Nonetheless, models trained on this data may amplify existing biases present in the social media platform. While this is in many cases unavoidable, we hope that by making this demo public with model prototypes, experts will be able to more easily inspect these biases and we will be able to better understand the potential biases of models trained on this type of data.

\section*{Acknowledgements}

We acknowledge the support of UKRI
(in particular the UKRI Future Leaders Fellowship scheme), Snap Inc., the Cardiff University Innovation for All scheme and the R\&D\&I grant PID2020-116118GA-I00 funded by MCIN/AEI/10.13039/501100011033 for partially funding this project.




\bibliography{anthology,custom}
\bibliographystyle{acl_natbib}

\appendix

\section{Languages supported}
\label{appendix:languages}

In addition to English, the sentiment analysis demo (including hashtag analysis) is also available for the following languages: Amharic, Arabic, Armenian, Basque, Bengali, Bulgarian, Burmese, Catalan, Chinese, Czech, Danish, Dhivehi, Dutch, Estonian, Finnish, French, Georgian, German, Greek, Haitian, Hebrew, Hindi, Hungarian, Icelandic, Indonesian, Italian, Japanese, Kannada, Khmer, Korean, Kurdish, Lao, Latvian, Lithuanian, Malayalam, Marathi, Nepali, Norwegian, Oriya, Panjabi, Persian, Polish, Pushto, Romanian, Russian, Serbian, Sindhi, Sinhala, Slovenian, Spanish, Swedish, Tagalog, Tamil, Telegu, Thai, Turkish, Uighur, Ukranian, Urdu, Vietnamese, Welsh. These languages are supported both by a XLM-T multilingual model and the Twitter API.








\section{Model Links}
\label{appendix:hf}

Table \ref{Table:model_links} lists all TweetNLP models and their corresponding Hugging Face model hub links.

\begin{table*}[!ht] 
\centering
\resizebox{1.0\textwidth}{!}{
\begin{tabular}{ll}
\toprule
\textbf{Model} & \textbf{Link} \\
\midrule
TweetEval & \url{https://huggingface.co/cardiffnlp/twitter-roberta-base} \\
TimeLMs-21 (default) & \url{https://huggingface.co/cardiffnlp/twitter-roberta-base-2021-124m} \\
XLM-T & \url{https://huggingface.co/cardiffnlp/twitter-xlm-roberta-base} \\
Sentiment Analysis & \url{https://huggingface.co/cardiffnlp/twitter-roberta-base-sentiment-latest} \\
Multilingual Sentiment Analysis & \url{https://huggingface.co/cardiffnlp/twitter-xlm-roberta-base-sentiment} \\
Emotion Recognition & \url{https://huggingface.co/cardiffnlp/twitter-roberta-base-emotion} \\
Emoji Prediction & \url{https://huggingface.co/cardiffnlp/twitter-roberta-base-emoji} \\
Irony Detection & \url{https://huggingface.co/cardiffnlp/twitter-roberta-base-irony} \\
Hate Speech Detection & \url{https://huggingface.co/cardiffnlp/twitter-roberta-base-hate} \\
Offensive Language Identification & \url{https://huggingface.co/cardiffnlp/twitter-roberta-base-offensive} \\
Stance Detection (abortion) & \url{https://huggingface.co/cardiffnlp/twitter-roberta-base-stance-abortion} \\
Topic Classification & \url{https://huggingface.co/cardiffnlp/tweet-topic-21-multi} \\
Named Entity Recognition & \url{https://huggingface.co/tner/twitter-roberta-base-dec2021-tweetner7-all} \\
Tweet Embeddings & \url{https://huggingface.co/cambridgeltl/tweet-roberta-base-embeddings-v1} \\

\bottomrule
\end{tabular}
}
\caption{Hugging Face model links of all the NLP models included in TweetNLP (if available).}
\label{Table:model_links}
\end{table*}


We release the word embeddings along with Gensim-optimized versions: (1) English-monolingual word embeddings are available at \url{https://tweetnlp.org/downloads/twitter-2021-124m-300d.new.bin}; (2) Multilingual word embeddings are available at \url{https://tweetnlp.org/downloads/twitter-multilingual-300d.new.bin}.


Table \ref{Table:baselinemodel_links} lists the baselines used for the evaluation (Section \ref{sec:eval}) and their corresponding Hugging Face hub links.

\begin{table*}[!ht] 
\centering
\resizebox{.88\textwidth}{!}{
\begin{tabular}{ll}
\toprule
\textbf{Model} & \textbf{Link} \\
\midrule
RoBERTa-base & \url{https://huggingface.co/roberta-base} \\
XLM-R & \url{https://huggingface.co/xlm-roberta-base} \\

BERTweet & \url{https://huggingface.co/vinai/bertweet-base} \\
Sentence-BERT & \url{https://huggingface.co/sentence-transformers/bert-base-nli-mean-tokens} \\
all-mpnet-base-v2 & \url{https://huggingface.co/sentence-transformers/all-mpnet-base-v2} \\
Mirror-RoBERTa & \url{https://huggingface.co/cambridgeltl/mirror-roberta-base-sentence-drophead} \\
SimCSE-RoBERTa & \url{https://huggingface.co/princeton-nlp/unsup-simcse-roberta-base} \\
\bottomrule
\end{tabular}
}
\caption{Baseline models' Hugging Face links (if available).}
\label{Table:baselinemodel_links}
\end{table*}


\end{document}